\documentclass[runningheads]{llncs}
\usepackage{graphicx}

\usepackage{tikz}
\usepackage{comment}
\usepackage{amsmath,amssymb} 
\usepackage{color}

\usepackage{booktabs}
\usepackage{pifont}
\usepackage{multirow}
\usepackage{algorithm}
\usepackage{algorithmic}
\usepackage{makecell}
\usepackage[pagebackref]{hyperref}
\usepackage{subfigure}

\usepackage[accsupp]{axessibility}  

\begin{document}
\pagestyle{headings}
\mainmatter
\def\ECCVSubNumber{4305}  

\title{Dynamic Low-Resolution Distillation for Cost-Efficient End-to-End Text Spotting} 


\titlerunning{Dynamic Low-Resolution Distillation (DLD)}
%
\author{Ying Chen\inst{2}\textsuperscript{$\star$} \and
Liang Qiao\inst{1,2}\thanks{Authors contributed equally. $\dagger$ Corresponding Authors.} \and Zhanzhan Cheng\inst{2}  \and  Shiliang Pu\inst{2}\textsuperscript{$\dagger$}  \and \\ Yi Niu\inst{2} \and Xi Li\inst{1}\textsuperscript{$\dagger$}}
\authorrunning{Chen. et al.}
%
\institute{Zhejiang University, Hangzhou, China \and
Hikvision Research Institute, Hanzhou, China \\
\email{\{chenying30,chengzhanzhan,pushiliang.hri,niuyi\}@hikvision.com \\ \{qiaoliang, xilizju\}@zju.edu.cn} \\
}
\maketitle

\begin{abstract}
End-to-end text spotting has attached great attention recently due to its benefits on global optimization and high maintainability for real applications. However, the input scale has always been a tough trade-off since recognizing a small text instance usually requires enlarging the whole image, which brings high computational costs.
In this paper, to address this problem, we propose a novel cost-efficient Dynamic Low-resolution Distillation (DLD) text spotting framework, which aims to infer images in different small but recognizable resolutions and achieve a better balance between accuracy and efficiency.
Concretely, we adopt a resolution selector to dynamically decide the input resolutions for different images, which is constraint by both inference accuracy and computational cost. Another sequential knowledge distillation strategy is conducted on the text recognition branch, making the low-res input obtains comparable performance to a high-res image. The proposed method can be optimized end-to-end and adopted in any current text spotting framework to improve 
the practicability. Extensive experiments on several text spotting benchmarks show that the proposed method vastly improves the usability of low-res models. The code is available at \url{https://github.com/hikopensource/DAVAR-Lab-OCR/}.

\end{abstract}

\section{Introduction}

Research on scene text spotting has achieved great process and been successfully applied in many fields such as finance, education, transportation, \textit{etc}. The traditional process of text spotting is usually divided into two sub-tasks: text detection~\cite{zhou2017east,baek2019character,Wang2019Shape,DBLP:conf/aaai/LiaoWYCB20} and recognition\cite{CRNN,cheng2017focus,DBLP:conf/aaai/WanHCBY20,DBLP:conf/cvpr/FangXWM021}.
To reduce the error accumulation between two tasks and the maintenance cost, many works have been proposed in an end-to-end manner~\cite{li2017towards,bartz2017see,he2018end,sun2018textnet,liao2019mask,qin2019towards,DBLP:conf/aaai/QiaoTCXNPW20}. To further improve the real-time performance of the model, some works exquisitely design different geometric representations or lighter network architectures~\cite{liu2018fots,liu2020abcnet,DBLP:journals/corr/abs-2105-03620,DBLP:journals/corr/abs-2105-00405,DBLP:conf/aaai/WangZQLZLHLDS21}.
However, most works only report the results based on a fixed and carefully selected input resolution, but whose performances are usually seriously affected by the resolution changing in different situations.

\begin{figure}[t]
\begin{center}
\includegraphics[width=0.9\linewidth]{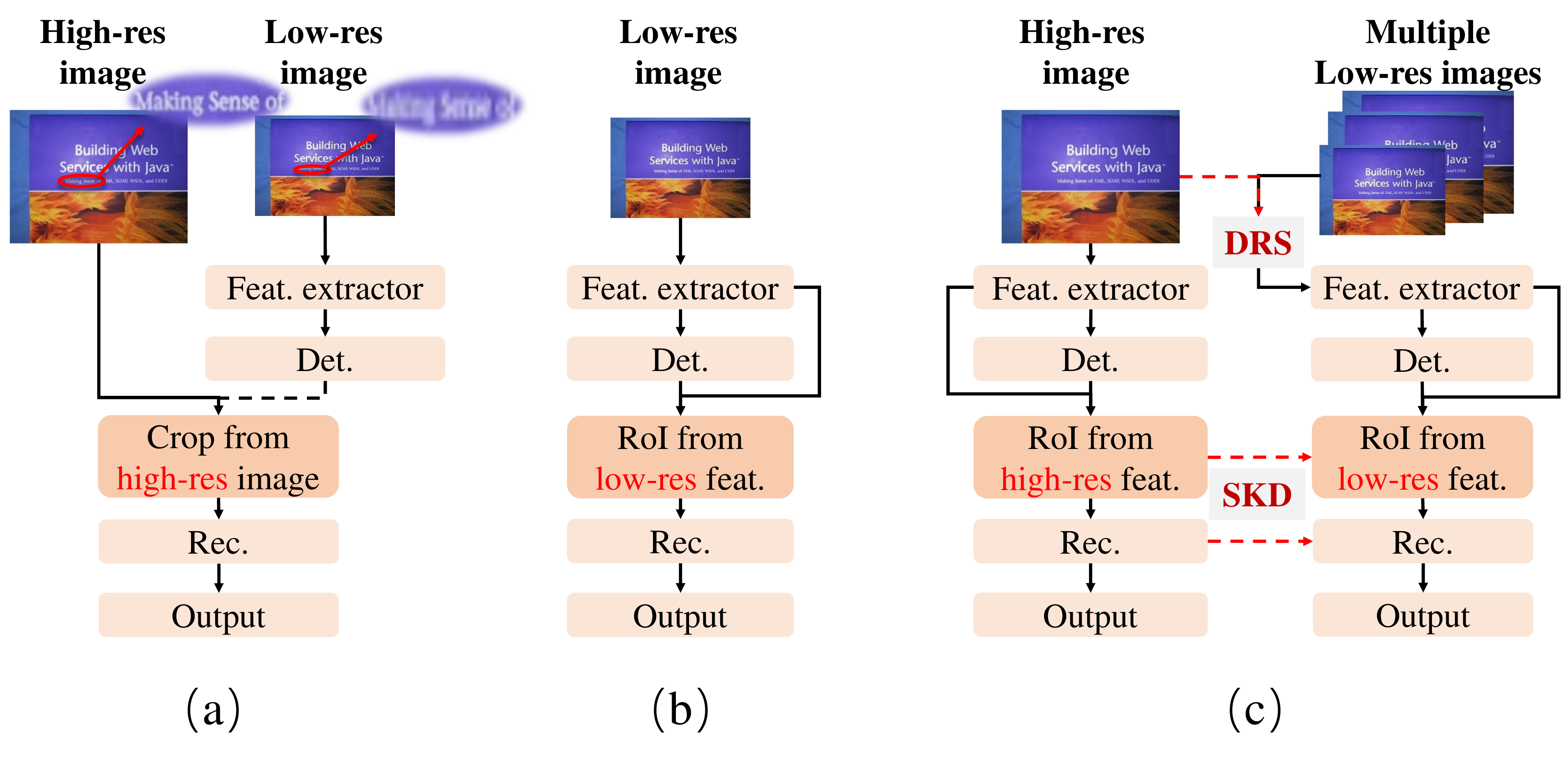}\\
\end{center}
\caption{
(a) is the offline two-staged text spotter, which can use different resolutions for two tasks while cannot be globally optimized. (b) is the ordinary end-to-end text spotter, where the recognizer can only receive the low-res RoI features map when using the low-res input. (c) is our proposed DLD framework, where the low-res network can dynamically select small but feasible resolutions and reconstruct the high-res features.
}
\label{fig:0}
\end{figure}
In the traditional pipeline of two-staged text spotting, as shown in Figure \ref{fig:0}(a), to save the inference cost, we can firstly detect text from a down-sampled image and then crop text regions from the original high-res image for recognition.
It will not damage the overall performance to some extent since the two tasks can be optimized separately.
However, once enjoying the benefits such as global optimization and lower maintenance cost brought by end-to-end text spotters, we have to face the resolution choice problem: images can only be resized into a predefined scale. If we want to achieve a higher efficiency using low-res inputs, as illustrated in Figure \ref{fig:0}(b), many small texts will lose their discriminative features from the beginning of the network and are hard to be recognized. This makes end-to-end text spotter low practicability in many realistic situations.

This problem is mainly attributed to the different characteristics and resolution sensitivities of two sub-tasks of end-to-end text spotters \cite{DBLP:journals/csur/ChenJZLW21}. When people read the text in a low-res image, they may easily identify whether an object belongs to a text. However, the blurred texture may influence the recognition since it is a more fine-grained sequential classification task. In fact, different texts may have different difficulties for recognition. People can sometimes correctly infer some of the low-res text according to the other recognizable characters and their semantic context meanings~\cite{DBLP:conf/cvpr/FangXWM021,DBLP:conf/cvpr/YuLZLHLD20,Bhunia_2021_Joint}. Nevertheless, when the image keeps down-sampling, it will gradually lose the distinct features for recognition. It means that for an image containing text with varied sizes and locations, a minimum resolution must exist to make all of the instances recognizable.

Therefore, a better way to balance accuracy and computational cost is to infer different images in different scales~\cite{DBLP:conf/cvpr/UzkentE20,DBLP:conf/cvpr/YangHCSDH20,DBLP:journals/corr/abs-2106-02898}. Moreover, in order to make the network prefer to choose the smaller scales with minimal accuracy drop, we borrow the idea of resolution Knowledge Distillation (KD)~\cite{DBLP:conf/eccv/YangZCYZW20,DBLP:conf/eccv/LiYC20,DBLP:conf/cvpr/QiKG0WCLJ21}, which can enhance the performance of the low-res student using the knowledge transferred from the high-res teacher.

In this paper, we propose a novel framework named \emph{Dynamic Low-resolution Distillation} (DLD) text spotting, which aims to
dynamically choose feasible input resolutions for the end-to-end text spotter under a resolution KD schema, as shown in Figure \ref{fig:0}(c).
Specifically, In DLD, the student network adopts a lightweight \emph{Dynamic Resolution Selector }(DRS) to find a suitable down-sampled resolution to preserve the teacher's performance. Given a group of candidate down-sampled scales, DRS is optimized to find the best resolution with a minimal performance drop under both accuracy and computational cost supervision.
On the other hand, to enhance the recognizability of low-res text, we emphasize the sequence information that the model extracts and integrate it with the \emph{Sequential Knowledge Distillation} (SKD) strategy in the recognition part. The loss of SKD is composed of feature-based L2 loss and the sequence-level beam search output loss, which effectively increases the performance of those \emph{low-res but recognizable} text instances.
DLD is a self-consistent framework, where the proposed two tasks can work together organically and achieve effective mutual promotion. The SKD prompts the DRS to choose a relatively lower resolution, and the DRS provides varied ratios of resolution pairs to enhance the SKD to be more robust and focus on the scale-unrelated features.

The major contributions of this paper are as follows:
(1) We first study the input resolution problem on end-to-end text spotting tasks and propose a Dynamic Low-resolution Distillation text spotting framework that can effectively enhance the performance and reduce the computational cost.
(2) We propose a sequential KD strategy with a dynamic resolution selector that allows the model to choose a small but recognizable input scale.
(3) Extensive experiments and ablation studies demonstrate the effectiveness of our method.

\section{Related Works}

\subsection{End-to-End Text Spotting}

Whether to employ Region of Interests (RoI) operations, current end-to-end text spotters can be divided into two types: two-staged and one-staged models.

Two-staged end-to-end text spotters usually involve RoI-like operations to crop detected regions from feature maps for the following recognition task.
Methods~\cite{DBLP:conf/iccv/LiWS17,liao2019mask,liao2020mask,lyu2018mask,qin2019towards,DBLP:conf/mm/ZhangXCP0QNW20} usually follow the Faster-RCNN~\cite{ren2015faster}/Mask-RCNN framework~\cite{he2017mask} to detect text regions and then crop the RoI regions into small features for recognition using RoI-pooling/RoI-Align operations. To recognize text in arbitrary shapes and improve detect efficiency, some works adopt the segmentation-based methods in the detection stage and then carefully design novel RoI operations to rectify the text into regular shapes, such as RoIRotate~\cite{liu2018fots}, RoI-Slide~\cite{feng2019textdragon}, TPS~\cite{DBLP:conf/aaai/QiaoTCXNPW20,DBLP:conf/aaai/WangLZYBXHW020}, BezierAlign~\cite{liu2020abcnet,DBLP:journals/corr/abs-2105-03620}, rectified RoI-Pooling~\cite{DBLP:conf/eccv/BaekSBPLNL20}, \textit{etc}.

In one-staged models, text instances are directly decoded from the global feature map without any RoI operation. \cite{xing2019convolutional} directly detects characters in different categories using multi-class segmentation. \cite{DBLP:conf/aaai/WangZQLZLHLDS21} decodes the gathered pixel-level feature vector into sequence with the proposed PG-CTC decoder. In the work of \cite{DBLP:conf/aaai/QiaoCCXNPW21}, the authors use mask attention to map different instances into different feature map channels and then predict individual text in each channel.

Both two types of methods suffer from the problem of input scales. Compared with two-staged methods that can reshape the RoI features into a uniform size, one-staged methods somehow face more challenges about text scales and usually require a large number of training samples.
\subsection{Knowledge Distillation}
Knowledge Distillation (KD) \cite{DBLP:journals/corr/HintonVD15} was first proposed to transfer the capacity of a large teacher network to a small student. This learning paradigm continues to evolve in the following years and has been applied in many areas, such as image recognition \cite{DBLP:conf/eccv/XuRLG20}, object detection \cite{DBLP:conf/cvpr/QiKG0WCLJ21}, semantic segmentation \cite{DBLP:conf/cvpr/HeSTGSY19}, text recognition \cite{Bhunia_2021_ICCV}.
In addition to transferring knowledge between different networks, resolution KD \cite{DBLP:conf/eccv/WangSLY20,DBLP:conf/eccv/YangZCYZW20,DBLP:conf/eccv/LiYC20} has also been widely used to train a low-res student with the help of a high-res teacher, which can be well adapted to improve some low-res applications like face recognition~\cite{DBLP:journals/tip/GeZLL19,DBLP:journals/tip/GeZLZL20}.
\cite{DBLP:conf/cvpr/QiKG0WCLJ21} firstly studies the low-res KD in object detection tasks and proposes an aligned multi-scale training method to align features in different levels.

However, these methods only conduct resolution KD in a fixed input resolution scale, which somehow limits the generalization ability of the model when objects are displayed in varied scales.

\subsection{Dynamic Resolution}
Since images have different difficulties to recognize, many works are proposed to assign images with dynamic input resolutions \cite{DBLP:conf/cvpr/UzkentE20,DBLP:conf/cvpr/YangHCSDH20,DBLP:conf/cvpr/WangKFYR19,DBLP:journals/corr/abs-2106-02898}.
\cite{DBLP:conf/cvpr/UzkentE20} proposes a reinforcement learning approach to dynamically identify when and where to use high-res data conditioned on the corresponding low-res data. \cite{DBLP:conf/cvpr/YangHCSDH20} uses different sub-networks to cope with samples with different difficulties. \cite{DBLP:conf/cvpr/WangKFYR19} learns how to use different scaling strategies for different objects. \cite{DBLP:journals/corr/abs-2106-02898} designs a resolution predictor to choose feasible input resolutions, which can reduce the computational cost while maintain performance.


\section{Methodology}
\subsection{Overview}
\begin{figure*}[t]
\begin{center}
\includegraphics[width=1.0\linewidth]{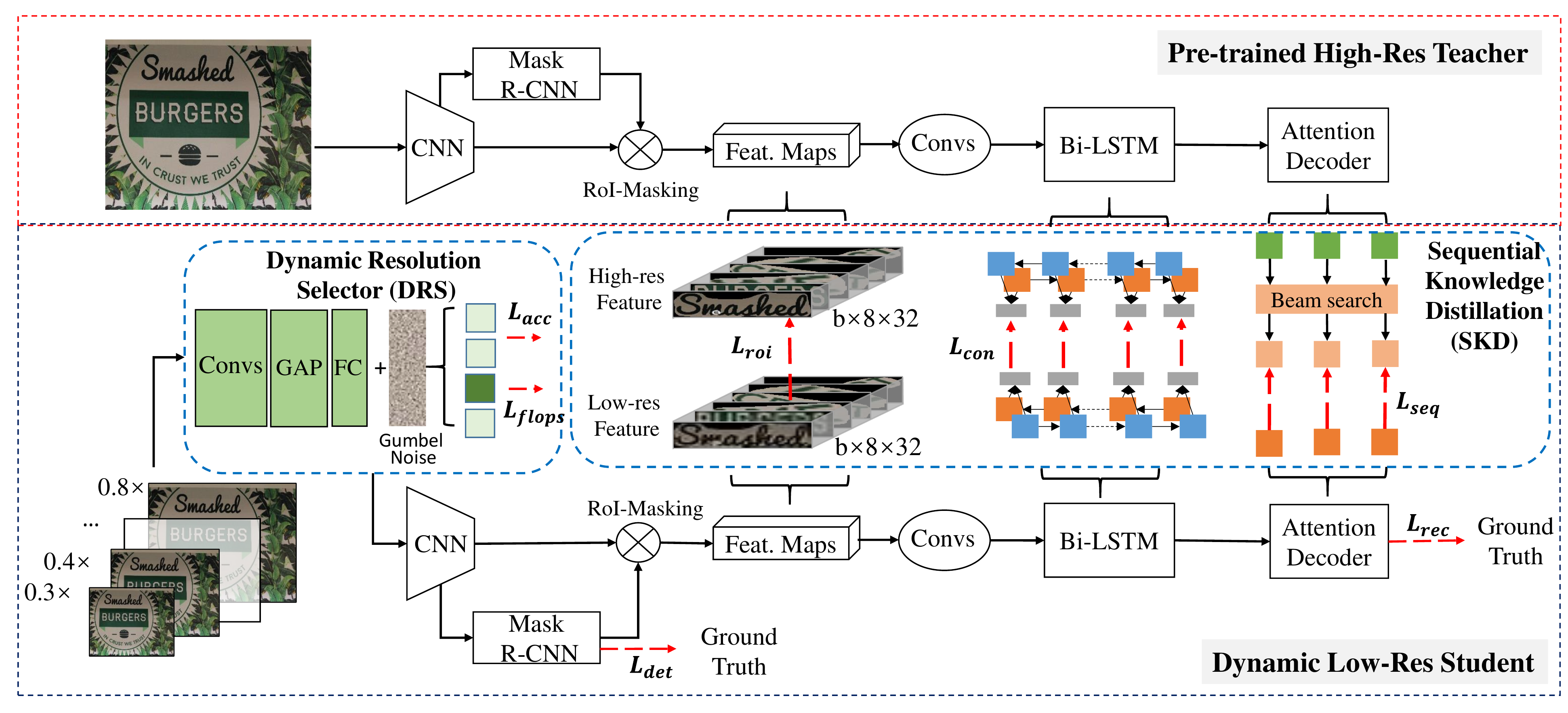}\\
\end{center}
\caption{
Illustration of the proposed DLD framework. It contains a fixed pre-trained high-res teacher and a dynamic low-res student that aims to obtain comparable performance. The DRS dynamically selects a small resolution for the student with minimal performance reduction. The SKD process helps the student capture the inter-sequence information and be able to recognize text in low-resolution.
}
\label{fig:1}
\end{figure*}
As shown in Figure~\ref{fig:1}, we propose a Dynamic Low-resolution Distillation (DLD) framework. It follows the setting of resolution distillation~\cite{DBLP:conf/eccv/WangSLY20,DBLP:conf/eccv/YangZCYZW20,DBLP:conf/eccv/LiYC20}, where the teacher and student use the same network architecture but were input with images in different resolutions. The high-res teacher network is well pre-trained and then fixed in the following training. The student aims to achieve comparable performance in some lower resolutions. It mainly attributes to two parts: (1) a Dynamic Resolution Selector (DRS) to choose the appropriate resolution for different input images, and (2) a Sequential Knowledge Distillation (SKD) strategy to capture the semantic sequence information and improve the recognizability of the low-res instances. The whole framework is end-to-end trainable.
\subsection{Baseline Text Spotting Model}
\label{sec:1}
We adopt a Mask-RCNN-based\cite{he2017mask} two-staged end-to-end text spotting framework~\cite{lyu2018mask,liao2019mask,qin2019towards,liao2020mask} as the base model. The detection branch follows the standard Mask-RCNN implementation, which can predict the mask region with its bounding box for any text shape.
For the input image $I$$\in$$R^{H\times W\times 3}$, the multi-scale features are extracted through the backbone of ResNet-50 \cite{he2016deep} and a feature pyramid network (FPN) \cite{lin2017feature}.
For the text recognition task, the recognition features are firstly cropped from the global feature map via the RoI-Masking operation\cite{qin2019towards} and then uniformly resized into a fixed size $H_{r}$$\times$$W_{r}$. $H_{r}$ and $W_{r}$ are set to 8 and 32 pixels separately in all experiments. These feature maps go through six convolution layers and are then extracted the sequence information by the bidirectional long short-term memory (Bi-LSTM)~\cite{hochreiter1997long} module. The final character sequences are decoded by an attention-based decoder~\cite{cheng2017focus}.

We firstly train a strong high-resolution teacher and then fix it in following distillation training. The teacher network are trained under the supervision from both text detection and recognition annotations as follows,
\begin{equation}
L_{teacher} = L_{det} + L_{rec},
\end{equation}
where the detection part contains the losses from bounding boxes regression, classification, and instance segmentation.
\subsection{Dynamic Resolution Selector}


Inspired by~\cite{DBLP:journals/corr/abs-2106-02898}, we propose a \emph{Dynamic Resolution Selector} (DRS) to help model inference images in more feasible scales.
Here, we predefine a group of candidate down-sampled scales for student network,
\textit{e.g.}, images can be selected in the range $[0.8\times, 0.3\times]$ of teacher's resolution.
The target of DRS is to find a suitable scale in the group with minimal performance reduction compared to the teacher. The selection criteria are that decreasing the resolution should be rewarded while performance decline would be penalized.
Specifically, The DRS is a lightweight residual network composed of 10 Convolution layers (Convs), a Global Average Pooling (GAP) layer, and a Fully Connected (FC) layer. Given a high-res image $I$ and $k$ candidate down-sampled scale factors $\{s_1,s_2,...,s_k\}$, the DRS first predicts the probability vector $p$$=$$[p_1,p_2,...,p_k]$ by the network and then transforms $p$ into binary decisions $h$$=$$[h_1,h_2,...,h_k]$$\in$$\left\{0,1 \right\}^k$ indicating which scale factor to select. To optimize DRS, the network first does forward calculations based on the given pre-trained teacher with the high-res input $I$ and all of the corresponding low-res images $\{I_{s_1},I_{s_2},...,I_{s_k}\}$. We use $y$ and $\{y_{s_1},y_{s_2},...,y_{s_k}\}$ to denote the predicted probability distribution of teacher and students, respectively. The loss generated in terms of accuracy can be formed as:
\begin{equation}
L_{acc} =  \mathcal{\mbox{\emph{KL}}}(\sum\limits_{i=1}^{k}h_{i}y_{s_i},y),
\end{equation}
where $\mathcal{\mbox{\emph{KL($\cdot$)}}}$ is the KL divergence for the recognition results.
The recognition feature maps are cropped according to the detection ground truth (GT) during training to ensure the number of recognition results is consistent. The optimization target is to make one of the $h_i$ be 1, and the others are 0. The selected scale with the teacher's nearest recognition results and the highest detection accuracy will obtain the minimum loss.

To prevent the DRS module from converging to the maximum scale and encourage it to choose a smaller image as much as possible, we directly penalize it with its forward computation cost as follows:
\begin{equation}
L_{flops} = \sum\limits_{i=1}^{k}h_{i}T_{i},
\end{equation}
where $T_{i}$ is the forward FLOPS under input $I_{s_i}$. Since images containing different instances have different FLOPS, we use the pre-computed average value.

The overall supervision of DRS is as follows:
\begin{equation}
L_{DRS} = L_{acc} + \gamma L_{flops},
\end{equation}
where $\gamma$ is the parameter to balance the weight between accuracy and computational cost.

\noindent \textbf{Gumbel Softmax Trick.} Notice that the process of converting $p$ into one-hot $h$ is non-differentiable. Here, we adopt the Gumbel-Softmax sampling trick~\cite{DBLP:conf/iclr/JangGP17}. Specifically, we first add Gumbel noise $g_j$ to the discrete random variable $p_j$, and then draw discrete samples from the above distribution as:
\begin{equation}
\label{eq2}
h_{i}=\left\{
\begin{aligned}
1 & , \quad i=\mathop{\arg\max}\limits_{j}(log(p_j)+g_j)\\
0 & , \quad \mbox{otherwise}
\end{aligned}
\right.,
\end{equation}
where $g_j$$=$$-log(-log(u_j))$ is calculated based on the i.i.d samples $u_j$, where $u_j$$\sim$$\mbox{\emph{Uniform}}(0,1)$.
In the above procedure, the $\mathop{\arg\max}$ operation can be approximated by softmax operation as follows:
\begin{equation}
h_i = \frac{exp((log(p_i)+g_i)/\tau)}{\sum\limits_{j=1}^{k}exp((log(p_j)+g_j)/\tau)},
\end{equation}
where $\tau$ is the temperature parameter. During training, using a lower $\tau$ can make the expectation of sampling closer to the result of $\mathop{\arg\max}$, but the gradient variance will be large. Adopting a higher $\tau$ can make the gradient variance smaller, but the expectation of sampling will be close to the average distribution. Here, we initialize with a larger $\tau$ at the beginning and gradually decrease it as $\tau$$=$$\sigma^{epoch}\tau_{init}$,
where $\tau_{init}$ is the initial temperature and $\sigma$ is a decay factor.
\subsection{Sequential Knowledge Distillation}

\begin{figure}[t]
\begin{center}
\includegraphics[width=0.9\linewidth]{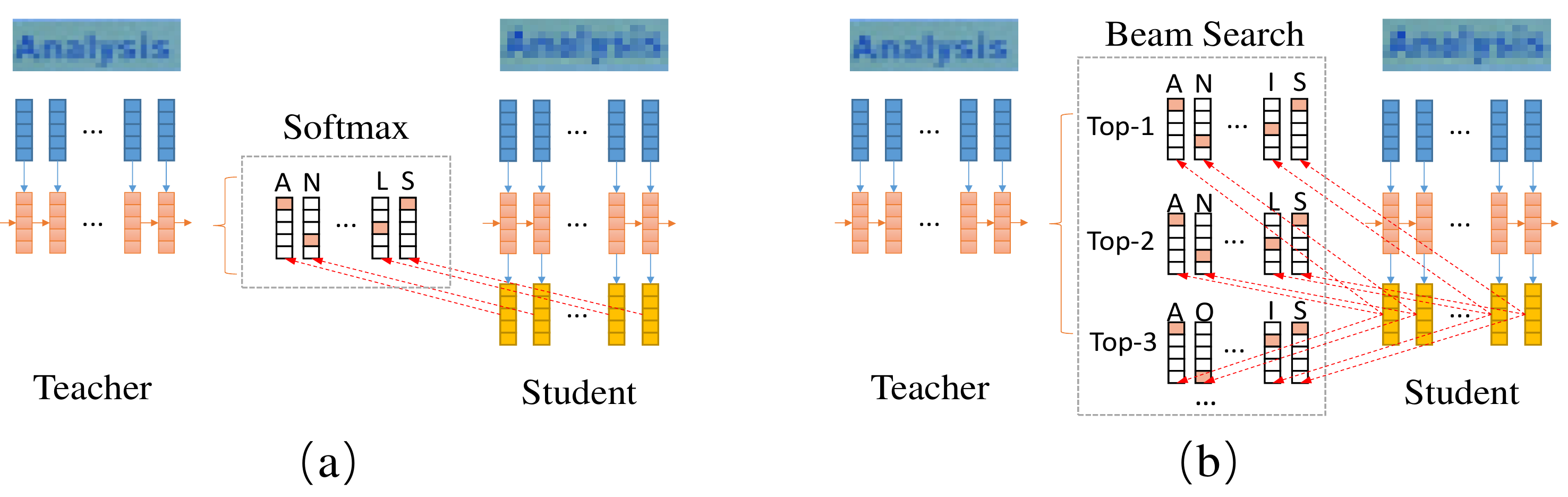}\\
\end{center}
\caption{
The comparison of (a) logit-based Knowledge Distillation and (b) sequence-level Knowledge Distillation in sequence decoding.
}
\label{fig:2}
\end{figure}

Text recognition is a sequential classification problem where the sequence information is vital for capturing the semantic meaning~\cite{CRNN,DBLP:conf/cvpr/FangXWM021,DBLP:conf/cvpr/YuLZLHLD20}. For example, although some characters are easily confused in low-res, 
such as `i' and `l' in Figure~\ref{fig:2}, 
people can still recognize them in a word.  This inspires us to dig out further the model's deeper potential, making DRS choose the smaller scale as much as possible.
Therefore, we propose a \emph{Sequential Knowledge Distillation} (SKD) strategy to help the low-res network extract semantic information under the supervision of its teacher.
Here, we only explore the optimization problem of the text recognition task since it is more likely to be the bottleneck of the overall performance of the current end-to-end text spotting framework, which will also be demonstrated in the following experiments.

Specifically, given a high-res input $I$ and the selected low-res image $I_{s}$, we use  $\mathcal{F}^{roi},\mathcal{F}^{roi}_{s}$$\in$$\mathbb{R}^{H\times W \times C}$ to separately represent the cropped recognition features of teacher and student, where $H$, $W$, and $C$ denote the feature map's height, width, and channel, respectively. These features are cropped using RoI-Masking with the detection GT and resized into a uniform shape. After a stack of convolutions and a Bi-LSTM module, the contextual information can be further extracted, which are represented as $\mathcal{F}^{con},\mathcal{F}^{con}_{s}$$\in$$\mathbb{R}^{N\times C}$, where $N$ is the length of the hidden state.
In these two stages of the network, we adopt the feature-based KD strategy to set up L2 loss as:
\begin{equation}
L_{roi} = \frac{1}{HWC}\sum\limits_{i=1}^{H}\sum\limits_{j=1}^{W}\sum\limits_{c=1}^{C}||\mathcal{F}^{roi}[i,j,c]-\mathcal{F}^{roi}_{s}[i,j,c]||_{2},
\end{equation}
\begin{equation}
L_{con} = \frac{1}{NC}\sum\limits_{i=1}^{N}\sum\limits_{c=1}^{C}||\mathcal{F}^{con}[i,c]-\mathcal{F}^{con}_{s}[i,c]||_{2}.
\end{equation}

In the final decoding stage, different from the distillation process that  aggregates logit-based loss~\cite{Bhunia_2021_ICCV} over the sequence, we borrow the idea of the sequence-level knowledge distillation~\cite{DBLP:conf/emnlp/KimR16} to better capture sequence information, as shown in Figure~\ref{fig:2}.  
The student is trained based on the output from the top-$k$ beam search~\cite{Rabiner1989A} results of the teacher network. It helps the student retain the context information in a sequence as much as possible. To ensure the optimization speed, we only choose the results with the top-$3$ scores.
Specifically, given the input $\mathcal{F}^{con}$ and the attention-based decoder, we use $p(q|\mathcal{F}^{con})$ to represent the predicted sequence distribution over all possible sequences $q$$\in$$\mathcal{Q}$.
Then, the sequence-level knowledge distillation can be formulated and then approximated as follows:
\begin{equation}
\begin{split}
L_{seq}
& = -\sum\limits_{q\in \mathcal{Q} }p(q|\mathcal{F}^{con})\log p(q|\mathcal{F}^{con}_{s}) \\
& = -\sum\limits_{q\in \mathcal{Q}}\mathbb{I}\{q=\hat{y}_1,\hat{y}_2,\hat{y}_3,... \} \log p(q|\mathcal{F}^{con}_{s}) \\
& \approx -\log \sum_{k=1}^K p(q=\hat{y}_k|\mathcal{F}^{con}_{s})
\end{split},
\end{equation}
where $\hat{y}_k$ is the result with the top-$k$ beam search score of the teacher model. Here, we simply set K=1 to save training time.

Finally, the overall SKD is optimized as:
\begin{equation}
L_{SKD} = L_{roi} + \eta_{1} L_{con} + \eta_{2} L_{seq},
\end{equation}
where $\eta_{1}$ and $\eta_{2}$ are the hyper-parameters to balance the magnitude of $L_{roi}$, $L_{con}$ and $L_{seq}$.

\subsection{Optimization}

The proposed DLD framework is optimized to recognize images in both high accuracy and efficiency with end-to-end training. The overall loss of the student network is generated from three parts: the original text detection and recognition loss, the loss from DRS to balance accuracy and computational cost, and the loss from SKD to improve the representational capacity of the low-res model.
\begin{equation}
L = L_{det} + \lambda_{1} L_{rec} +\lambda_{2}  L_{DRS} + \lambda_{3} L_{SKD},
\end{equation}
where $\lambda_{1}$,$\lambda_{2}$ and $\lambda_{3}$ are weight balancing parameters.

In every epoch of the training stage, the student network will do forward calculations for $k$ times for all candidate resolutions. The backward loss of $L_{DRS}$ will only be propagated in the lightweight DRS module. The loss of $L_{det}$, $L_{rec}$, and $L_{SKD}$ will only be conducted on the branch with the maximum $h_{i}$ and will be not propagated to the DRS.
\section{Experiments}
\subsection{Implementation Details}
\noindent \textbf{Datasets.} We list the datasets used in this paper as follows. We evaluate our method on three popular text spotting benchmarks: (1) \emph{ICDAR2013}~\cite{karatzas2013icdar} (\emph{IC13}) that only contains horizontal text, (2) \emph{ICDAR2015} \cite{karatzas2015icdar} (\emph{IC15}) that includes oriented text, and (3) \emph{Total-Text}~\cite{ch2017total} (\emph{TT}) that involves many curved text.
For the teacher network, we firstly pre-train it on \emph{SynthText-800K} \cite{gupta2016synthetic} and then fine-tune with a mixture dataset which includes 7k images filtered from \emph{ICDAR-MLT2017} \cite{DBLP:conf/icdar/NayefYBCFKLPRCK17} and all training images in \emph{IC13}, \emph{IC15}, and \emph{TT}.
In the following KD training stage, the teacher will be fixed, and the student network can be initialized using the teacher's weights.

\noindent \textbf{Experiment Settings.}
The base model's architecture is described in Section~\ref{sec:1}. Teacher and student models share the same training settings. All models are trained by the AdamW~\cite{DBLP:conf/iclr/LoshchilovH19} optimizer with batch-size=$3$. The KD training lasts for $50$ epochs and uses an initial learning rate of $1$$\times$$10^{-3}$. The learning rate is divided by $10$ at the $30$-th
epoch and the $40$-th epoch. The parameter $\tau_{init}$ is set as $5$ and decay factor $\sigma$$=$$0.965$. The Bi-LSTM module has 256 hidden units. For weight balancing parameters, we set $\gamma$$=$$0.1$ and others as  $\eta_1$$=$$\eta_2$$=$$\lambda_1$$=$$\lambda_2$$=$$\lambda_3$$=$$1$.

According to the scales of text instances in different datasets, we choose the basic inference resolutions for the teacher, \emph{i.e.}, `S-768' for IC13, `S-1280' for IC15, and `S-896' for Total-Text, where the prefix of `\emph{S-}' represents the input images are resized by a fixed shorter side.

To obtain a strong baseline of the teacher network, we conduct widely-used data augmentation strategies as follows: (1) instance aware random cropping, (2) randomly scaling
the shorter side of the input images to lengths in the range scales [$0.3$$\times$, $1.0$$\times$] of the basic resolutions, (3) random rotation with angle randomly chosen
from $[-15^{\circ}, +15^{\circ}]$, (4) applying random brightness, jitters, and contrast to input images.
In both training and testing stages, the student's DRS resolution ratios range is set as $\{0.8, 0.7, 0.6, 0.5, 0.4, 0.3\}$.
All experiments are implemented in Pytorch with 8$\times$32 GB-Tesla-V100 GPUs under CUDA-10.0 and CUDNN-7.6.3.
\subsection{Results on Text Spotting Benchmarks}
The effectiveness of the proposed DLD are compared with other three settings:
(1) \emph{Vanilla Multi-scale}: a single model that is trained in multiple scales and tested in a fixed scale.
(2) \emph{DRS-only}: with the proposed DRS, the student network removes the supervision from the distillation.
(3) \emph{SKD-only}: with the proposed SKD, the student keeps using the $1/2$ scale compared with the teacher.

\begin{table}[ht]
\begin{minipage}[t]{1\linewidth}
\caption{Results on three text spotting benchmarks. `S', `W' and `G' separately mean recognition with strong, weak and generic lexicon~\cite{karatzas2015icdar}. `Full' indicates lexicons of all images are combined, and `None' means lexicon-free~\cite{ch2017total}. `H' and `L' in column `Type' indicates whether the inference is carried out with high- or low-resolution input.  FLOPS is the average \emph{floating point operations}.
}
\begin{center}
\scalebox{0.8}{
\begin{tabular}{c|l|c|c|c|c|c|c|c|c|c|c|c|c|c}
\hline
\multirow{2}{*}{Dataset} & \multirow{2}{*}{Training Method}  & \multirow{2}{*}{Type} &\multirow{2}{*}{Input Size} & \multicolumn{5}{c|}{End-to-End (\%)} & \multicolumn{5}{c|}{Word Spotting (\%)} & \multirow{2}{*}{FLOPS} \\ \cline{5-14}
 & & & & S & W & G & None & Full & S & W & G & None & Full &  \\
\hline
\multirow{7}{*}{IC13}
& \multirow{2}{*}{Vanilla Multi-Scale}  & H & S-768 & 86.9 & 86.6 & 82.9 & -& - &91.4 & 91.0 & 86.3 & -& - & 142.9G \\
&  & L & S-384 & 80.9 & 78.9 & 74.4 & -& - & 85.2 & 82.7 & 77.3 & -& - & 35.8G \\
\cline{2-15}
& SKD-only & L & S-384  &  84.1 & 82.8 & 78.8 & -& - & 88.0 & 86.5 & 81.7 & -& - &35.8G\\
\cline{2-15}
& DRS-only ($\gamma$$=$$0.1$)  & L & Dynamic  & 85.7 & 84.8 & 80.7 &-& - & 90.1 & 88.9 & 84.0 &-& - & 80.7G \\
& DRS-only ($\gamma$$=$$0.3$)  & L & Dynamic  & 83.7 & 82.0 & 77.6 &-& - & 87.8 & 85.8 & 80.5 & -& - &48.8G\\
\cline{2-15}
& {DLD} ($\gamma$$=$$0.1$)  & L & Dynamic &  86.5 & 85.7 & 82.7 & -& - & 90.9 & 89.9 & 86.1 & -& - &71.5G\\
& {DLD} ($\gamma$$=$$0.3$)  & L & Dynamic &  85.6 & 84.4 & 81.6 & -& - & 90.0 & 88.6 & 84.9 & -& - &41.6G\\
\hline
\multirow{7}{*}{IC15}
& \multirow{2}{*}{Vanilla Multi-Scale}  & H & S-1280 & 78.0 & 74.4 & 69.5 &-& - & 81.4 & 77.2 & 71.7 &  -& - &517.2G \\
&  & L & S-640 & 72.2 & 67.8 & 62.9 &-& - & 75.7 & 70.8 & 65.3 &  -& - &129.3G \\
\cline{2-15}
& SKD-only & L & S-640  & 75.4 & 71.7 & 67.1 &-& - & 78.9 & 74.6 & 69.6 & -& - &129.3G\\
\cline{2-15}
& DRS-only ($\gamma$$=$$0.1$)  & L & Dynamic  & 76.2 & 72.1 & 66.8 &-& - & 79.8 & 75.2 & 69.3 &-& - & 298.8G \\
& DRS-only ($\gamma$$=$$0.3$)  & L & Dynamic  & 73.6 & 68.9 & 63.7 &-& - & 76.4 & 71.5 & 66.3 & -& - &163.6G\\
\cline{2-15}
& {DLD ($\gamma$$=$$0.1$)}  & L & Dynamic &  79.0 & 75.7 & 70.9 &-& - & 82.4 & 78.6 & 73.3 & -& - &261.8G \\
& {DLD ($\gamma$$=$$0.3$) }  & L & Dynamic &  78.1 & 73.5 & 69.3 &-& - & 81.1 & 76.4 & 71.2 & -& - &148.3G \\
\hline
\multirow{7}{*}{TT}
& \multirow{2}{*}{Vanilla Multi-Scale}  & H & S-896 & - & - & -   & 62.3  & 71.4& - &  - & - & 65.2 &75.9 & 206.7G \\
&  & L & S-448 & - & - & -   & 55.4 & 66.5 & - &  - & - & 58.1 &71.1 & 52.0G \\
\cline{2-15}
& SKD-only   & L & S-448  & - & - & -   & 59.6 & 68.9 & - &  - & -  & 62.6 & 73.5 &52.0G \\
\cline{2-15}
& {DRS-only ($\gamma$$=$$0.1$)}  & L & Dynamic   & - & - & -   & 60.9 & 70.4 & - &  - & -  & 63.5 & 75.0 &119.2G \\
& {DRS-only ($\gamma$$=$$0.3$)}  & L & Dynamic   & - & - & -   & 58.8 & 68.9 & - &  - & -  & 61.6 & 73.6 &75.0G \\
\cline{2-15}
& {DLD ($\gamma$$=$$0.1$)}  & L & Dynamic   & - & - & -   & 63.9 & 73.7 & - &  - & -  & 66.4 & 77.8 &103.0G \\
& {DLD ($\gamma$$=$$0.3$)}  & L & Dynamic   & - & - & -   & 61.9 & 71.9 & - &  - & -  & 64.0 & 75.9 &62.1G \\
\hline
\end{tabular}
}
\end{center}
\label{tb:1}
\end{minipage}
\end{table}

Table \ref{tb:1} shows the experimental results on three benchmarks. The result of \emph{Vanilla Multi-scale} with high-res input can be treated as the original upper bound of the model. If the inputs are in low-resolutions, although the FLOPS can be optimized by approximately 75\%, the accuracies will be dramatically decreased, \textit{e.g.}, the \emph{End-to-End} results of \emph{General/None} are decreased by 8.5\% (82.9\% $vs.$ 74.4\%) in \emph{IC13}, 6.6\% (69.5\% $vs.$ 62.9\%) in \emph{IC15} and 6.9\% (62.3\% $vs.$ 55.4\%) in \emph{Total-Text}, respectively.
Using the SKD to transfer knowledge from the high-res teacher into the low-res student, in the result of \emph{SKD-only}, we can see that the performances on low-res can be effectively increased by 4.4\%/4.2\%/4.2\% in three datasets compared with \emph{Vanilla Multiscale}, respectively.

Equipped with the DRS module, in \emph{DRS-only}, we are able to tune the tendencies of the model to balance the accuracy and computational cost by different $\gamma$. When the model additionally integrates with SKD, under the entire \emph{DLD} framework, where the student network can select more low-res scales without performance drop, the overall performances are further optimized. Specifically, when we set $\gamma$=$0.1$, the model can achieve the comparable or even better accuracy (82.7\% $vs.$ 82.9\%, 70.9\% $vs.$ 69.5\%, 63.9\% $vs.$ 62.3\%) than that of high-res input, and with about 50\% FLOPS costs. Suppose we want the model more tend to be cost-efficient and set $\gamma$=$0.3$. The models' FLOPS can be reduced to a similar level as that all using $1/2$-resolution inputs, and the accuracies can be further improved by 2.8\%/2.2\%/2.3\% compared with \emph{SKD-only}, respectively. More statistical and visualized analysis are in the supplementary materials.

We conduct the following ablation experiments in Section \ref{sec:4.3},\ref{sec:4.4},\ref{sec:2} on \emph{Total-Text}~\cite{ch2017total} and use `None' to denote lexicon-free End-to-End result and `Full' to represent the result based on the lexicon combined all images.

\subsection{Ablation Studies on Sequential Knowledge Distillation}
\label{sec:4.3}

\noindent \textbf{Different Distillation Losses.} 
SKD contains losses from three parts: the RoI feature's loss $L_{roi}$, the contexture feature's loss $L_{con}$, and the beam search output's loss $L_{seq}$.
Here, we conduct different experiments based on the model of \emph{SKD-only} to evaluate the importance degrees of these losses, and the result is shown in Table~\ref{tb:3}. It is easy to know that without any KD loss, the model will fall into the \emph{Vanilla Multi-Scale} setting. From the result, we can see that $L_{seq}$ has the greatest impact on distillation, which surpasses the result without distillation by 3.0\% on `None' and 1.4\% on `Full'. By combining all three losses, the model achieves 4.2\%/2.4\% improvements.

\noindent \textbf{Different Knowledge Distillation Settings.}
We also conduct experiments on different KD strategies. Based on the setting of Resolution KD framework without DRS (teacher with S-896, and student with S-448), we compared our model with other two works: (1) Bhunia \emph{et al.}~\cite{Bhunia_2021_ICCV}: contains four types of KD losses (Logits' Distillation, Character Localised Hint, Attention Distillation, Affinity Distillation) that are designed for text recognition, and (2) Qi \emph{et al.}~\cite{DBLP:conf/cvpr/QiKG0WCLJ21}: a KD strategy designed for the detection stage.
Table~\ref{tb:4} shows the experimental results. For the recognition KD, the results demonstrate that the performance of our proposed SKD surpasses~\cite{Bhunia_2021_ICCV} by 0.4\%/0.6\%. This is mainly because of the effectiveness of the sequence-level distillation strategy in some low-res texts. We simply replace the sequence-level distillation in SKD with the logits-based distillation adopted in \cite{Bhunia_2021_ICCV}, and we can see the performance will drop by 0.9\% on `None' and 1.4\% on `Full', respectively.
For the detection KD, the results show that the enhancement of integration with detection distillation is limited. This is because, in the current model, the detection performance in low-resolution has only a small gap between that of high-resolution.

\noindent \textbf{Different RoI Scales.} Recognition feature scale is a factor that affects performance under the current Mask-RCNN-based framework. Here, we conduct experiments to evaluate its influence, whose results are illustrated in Table \ref{tb:4-2}. The results show that larger RoI scales would help models obtain higher performance but inevitably bring extra computational cost.
On the other hand, if teacher and student use different RoI scales, the model cannot directly perform distillation. So, we add a deconvolutional (deconv) module \cite{DBLP:conf/cvpr/ZeilerKTF10} to align the smaller student's features with the larger teacher's features. This also simulates a feature-level super-resolution process. From the results, we can see that compared with directly using a larger student RoI scale, conducting super-resolution will even degrade the performance and increase the FLOPS.

\begin{table}[t]
\begin{minipage}[t]{0.48\linewidth}
\caption{Ablation on different distillation losses for SKD.}
\begin{center}
\scalebox{0.8}{
\begin{tabular}{c|ccc|c|c}
\hline
\multirow{2}{*}{\makecell[c]{Training \\Method}}  & \multirow{2}{*}{$L_{roi}$}  & \multirow{2}{*}{$L_{con}$} &\multirow{2}{*}{$L_{seq}$} & \multicolumn{2}{c}{End-to-End(\%)} \\ \cline{5-6}
& & & & None & Full   \\
\hline

\multirow{5}{*}{\makecell[c]{SKD-only \\ (S-448)}}& & &  & 55.4 & 66.5  \\
 & \ding{51}   &  &  & 56.9 & 66.9\\
 &  & \ding{51} &  & 57.5 & 67.2\\
 &  &  & \ding{51} & 58.4 & 67.9\\
 & \ding{51} & \ding{51} & \ding{51} & 59.6 & 68.9  \\
\hline
\end{tabular}
}
\end{center}
\label{tb:3}
\end{minipage}
\begin{minipage}[t]{0.1\linewidth}
\end{minipage}
\begin{minipage}[t]{0.5\linewidth}
\caption{Ablation on different knowledge distillation settings.}
\begin{center}
\scalebox{0.8}{
\begin{tabular}{c|c|c}
\hline
 \multirow{2}{*}{\makecell[c]{Distillation\\ Method}} & \multicolumn{2}{c}{End-to-End (\%))} \\ \cline{2-3}
&  None & Full   \\
\hline
 SKD & 59.6 & 68.9 \\
\hline
  Bhunia \emph{et al.}~\cite{Bhunia_2021_ICCV} & 59.2 & 68.3 \\
  SKD replaced Logits~\cite{Bhunia_2021_ICCV}& 58.7 & 67.5 \\
\hline
  Qi \emph{et al.}~\cite{DBLP:conf/cvpr/QiKG0WCLJ21} & 55.8 & 67.0 \\
  SKD+Qi \emph{et al.}\cite{DBLP:conf/cvpr/QiKG0WCLJ21} & 59.8 & 69.2 \\
\hline
\end{tabular}
}
\end{center}
\label{tb:4}
\end{minipage}
\begin{minipage}[t]{1\linewidth}
\caption{Ablation on different RoI scales. `$\dag$' means model has extra deconv modules.}
\begin{center}
\scalebox{0.8}{
\begin{tabular}{c|c|c|c|c|c|c}
\hline
\multirow{2}{*}{\makecell[c]{Training\\ Method}} & \multirow{2}{*}{\makecell[c]{Type}} & \multirow{2}{*}{\makecell[c]{Teacher \\ RoI Scale }} & \multirow{2}{*}{\makecell[c]{Student \\ RoI Scale }}  &\multicolumn{2}{c|}{End-to-End (\%))} &  \multirow{2}{*}{\makecell[c]{FLOPS }} \\ \cline{5-6}
& & & & None & Full &   \\
\hline
\multirow{2}{*}{Vanilla Multi-Scale} & H & (8$\times$32) & - & 62.3 & 71.4 & 206.7G \\
 & H &(16$\times$64) & - & 63.1 & 72.1 & 227.3G \\
\hline
\multirow{3}{*}{DLD} &  L &(8$\times$32) & (8$\times$32) & 63.9 & 73.7 & 103.0G \\
  & L & (16$\times$64)  &(16$\times$64) & 64.8 & 74.1 & 112.3G \\
   & L & (16$\times$64)  &(8$\times$32)$\dag$ & 64.5 & 73.3 & 119.2G \\
\hline
\end{tabular}
}
\end{center}
\label{tb:4-2}
\end{minipage}
\end{table}

\subsection{Ablation on Dynamic Resolution Selector }
\label{sec:4.4}

\noindent \textbf{Different Candidate Scales.} The set of candidate student scales is usually defined by experience. In Table~\ref{tb:5-1}, we compared the results under different candidate sets. The sets containing a single value are the same as the \emph{SKD-only} model. We can see that the group of smaller candidates \{0.5,0.4,0.3\} obtains lower inference costs and accuracy than that of the larger group \{0.8,0.7,0.6\}. With more candidates values, the model can be optimized to find a better balance between accuracy and computational cost. Nevertheless, the training cost will be somehow increased.
\begin{table}[t]
    \begin{minipage}[t]{1\linewidth}
    \caption{Ablation on different candidate scales. The training time reports the average time used to train the model for an epoch.  }
    \centering
    \scalebox{0.8}{
   \begin{tabular}{c|c|c|c|c|c}
        \hline
        \multirow{2}{*}{\makecell[c]{Training\\ Method}} &\multirow{2}{*}{Resolution Scales}  & \multicolumn{2}{c|}{End-to-End (\%)}  & \multirow{2}{*}{FLOPS} & \multirow{2}{*}{\makecell[c]{Training time\\ (min/epoch)} } \\ \cline{3-4}
         & & None & Full  &  & \\
        \hline
       \multirow{4}{*}{DLD} & \{0.5\} & 59.6 & 68.9 & 52.0G & 7.8\\
        &\{0.8, 0.7, 0.6\} & 63.6 & 73.4 & 128.8G & 9.0\\
        &\{0.5, 0.4, 0.3\} & 58.9 & 68.0 & 45.9G & 8.4 \\
        &\{0.8, 0.7, 0.6, 0.5, 0.4, 0.3\} & 63.9 & 73.7 & 103.0G & 10.2 \\
        \hline
        \end{tabular}
        }
        \label{tb:5-1}
    \end{minipage}

    \begin{minipage}[t]{0.49\linewidth}
    	\caption{Ablation on parameter $\gamma$. }
		\centering
        \scalebox{0.8}{
		\begin{tabular}{c|c|c|c|c}
        \hline
        \multirow{2}{*}{\makecell[c]{Training\\ Method}} & \multirow{2}{*}{$\gamma$}  & \multicolumn{2}{c|}{End-to-End (\%)}  & \multirow{2}{*}{FLOPS}\\ \cline{3-4}
        & & None & Full  &  \\
        \hline
        \multirow{5}{*}{DLD}& 0.1 & 63.9 & 73.7 & 103.0G\\
        & 0.2 & 63.2 & 72.4 & 82.8G\\		
        & 0.3 & 61.9 & 71.9 & 62.1G\\
        & 0.4 & 59.2 & 69.5 & 50.6G\\		
        & 0.5 & 56.0 & 66.4 & 38.4G\\
        \hline
        \end{tabular}
        }
		\label{tb:5-2}
	\end{minipage}
	\begin{minipage}[t]{0.49\linewidth}
\caption{Ablation on parameter $\tau_{init}$. }
		\centering
        \scalebox{0.8}{
		\begin{tabular}{c|c|c|c|c}
        \hline
       \multirow{2}{*}{\makecell[c]{Training\\ Method}} & \multirow{2}{*}{$\tau_{init}$}  & \multicolumn{2}{c|}{End-to-End (\%)}  & \multirow{2}{*}{FLOPS}\\ \cline{3-4}
        && None & Full  &  \\
        \hline
        \multirow{5}{*}{DLD}&  1 & 63.6 & 71.9 & 115.8G\\
        & 3 & 63.5 & 73.2 & 110.2G\\
        & 5 & 63.9 & 73.7 & 103.0G\\
        & 7 & 63.8 & 72.6 & 96.7G\\
        & 9 & 63.2 & 72.5 & 98.5G\\
        \hline
        \end{tabular}
        }
		\label{tb:5-3}
	\end{minipage}
\end{table}

\noindent \textbf{Balance between Accuracy and Computational Cost.}
$\gamma$ is a vital parameter to control the DRS's tendencies about accuracy and efficiency. Table~\ref{tb:5-2} shows how the changes of $\gamma$ influence the model. As the increasing of $\gamma$, the model tends to choose more minor input scales and achieves more efficient computational cost. However, the accuracy will be somehow reduced.
This parameter can be controlled flexibly and provide a straightforward guide on the resolution choice under different requirements.

\noindent \textbf{Different Temperature Parameter.} Table~\ref{tb:5-3} shows how $\tau_{init}$ influence the student's performance. This parameter affects the results of the convergence of the DRS module to some extent. It can be easily tuned once $\gamma$ be fixed.

\subsection{Studies on Different End-to-End Text Spotters}
\label{sec:2}
To demonstrate the effectiveness of our method, besides the basic Mask-RCNN-based text spotter, we also conducted experiments on the other two different frameworks based on the open-sourced code: (1) Text Perceptron \cite{DBLP:conf/aaai/QiaoTCXNPW20}, a two-staged text spotter whose text detection branch is segmentation-based, and (2) MANGO \cite{DBLP:conf/aaai/QiaoCCXNPW21}, a one-staged text spotter. Since MANGO has no explicit detection branch and directly recognizes text globally, we only report \emph{End-to-End} result based on Intersection over Union(IoU)$=$$0.1$ constraint as reported in \cite{DBLP:conf/aaai/QiaoCCXNPW21}. Other results are reported based on IoU$=$$0.5$. We calculate the \emph{Text Recognition (Rec)} accuracy using the GT of detection in inference.

The experimental results are displayed in Table~\ref{tb:6}. In the \emph{Vanilla Multi-Scale} setting, as the input scales decreases, it is not hard to understand that almost all of the accuracies will be declined. Nevertheless, in the Mask-RCNN-based framework, the performance drops in \emph{Rec} and \emph{End-to-End} are faster than that of \emph{Det}. It means the text detection task can still work well in some relatively low resolutions, while the text recognition task becomes the overall bottleneck of End-to-End performance. This also proves the motivation we mentioned initially. In contrast, for the other two text spotters, the \emph{Det} performance degrades faster when compressing the input size, which also jointly influences the End-to-End performance. This is because, in Mask-RCNN-based text spotters, there are a lot of preset anchors to capture different scales of text, but the segmentation task is relatively more sensitive to scales \cite{DBLP:conf/wacv/RichardsonAAGRA20}.
\begin{table}[t]
\caption{The ablations experiments on scale changes in \emph{Vanilla Multi-Scale} and the compared result when adopting DLD on different text spotting frameworks. `\emph{Det}' is the Hmean metric of text detection task. `\emph{Rec}' is the Accuracy metric of text recognition task. `\emph{E2E}' stands for `End-to-End'. `Full' indicates lexicons of all images are combined and `None' means lexicon-free. FPS is the average \emph{frames per second}.}
\begin{center}
\resizebox{1\textwidth}{!}{
\begin{tabular}{c|c|c|c|c|c|c|c|c|c|c|c|c|c|c}
\hline
\multirow{3}{*}{\makecell[c]{Training\\ Method}} & \multirow{3}{*}{\makecell[c]{Input\\ Size}} & \multicolumn{5}{c|}{Mask-RCNN-based}  & \multicolumn{5}{c|}{Text Perceptron \cite{DBLP:conf/aaai/QiaoTCXNPW20}}  &\multicolumn{3}{c}{MANGO \cite{DBLP:conf/aaai/QiaoCCXNPW21}} \\ \cline{3-15}
& & \multirow{2}{*}{\makecell[c]{Det\\(\%)}} & \multirow{2}{*}{\makecell[c]{Rec\\(\%)}} & \multicolumn{2}{c|}{E2E(\%)} & \multirow{2}{*}{FPS}& \multirow{2}{*}{\makecell[c]{Det\\(\%)}} & \multirow{2}{*}{\makecell[c]{Rec\\(\%)}} & \multicolumn{2}{c|}{E2E(\%)}&\multirow{2}{*}{FPS}& \multicolumn{2}{c|}{E2E(\%)}& \multirow{2}{*}{FPS} \\ \cline{5-6} \cline{10-11} \cline{13-14}
& & &  & None & Full &  &  & &None & Full &    & None & Full & \\
\hline
\multirow{6}{*}{\makecell[c]{Vanilla\\ Multi-Scale}} & S-896 & 85.3 &73.6 &62.3 & 71.4&7.9 &85.3 &73.1 & 66.0& 74.6& 9.5 & 66.2 & 77.7 & 3.7 \\
&S-768 & 85.7 &72.8 &61.2 & 71.1&8.6 &85.1 &72.6 & 64.7& 74.2& 11.4  & 67.1 & 78.1 & 4.7  \\
&S-640 & 86.1 &71.7 &60.5 & 70.2&9.0 &85.2 &72.2 & 64.3& 73.1& 13.3  & 66.5 & 76.5 & 6.1 \\
&S-512 & 84.9 &66.9 &58.7 & 69.3&9.4 &82.5 &68.9 & 61.3& 70.5& 14.8  & 61.9 & 73.3 & 7.7  \\
&S-384 & 82.3 &58.6 &52.0 & 63.5&9.8 &77.4 &60.7 & 54.1& 64.3& 16.5  & 50.9 & 64.4 & 11.5  \\
&S-256 & 76.5 &42.5 &38.3 & 48.5&12.0 &62.2 &45.2 & 40.5& 50.2& 17.8  & 29.2 & 46.3 & 13.9  \\
\hline
DLD ($\gamma$$=$$0.1$)  & Dynamic & 85.8 &74.8 &63.9 & 73.7&9.1 &85.6 &75.3 & 67.1& 76.4& 13.2  & 67.8 & 78.3 & 6.5  \\
DLD ($\gamma$$=$$0.3$) & Dynamic & 85.1 &73.1 &61.9 & 71.9&9.7 &81.7 &73.5 & 63.6& 72.8& 15.8  & 62.5 & 73.8 & 10.4  \\
\hline
\end{tabular}
}
\end{center}
\label{tb:6}
\end{table}

When we adopt the proposed DLD on these text spotters, we can see that all the models with $\gamma$$=$$0.1$ can obtain even higher accuracies and faster speed than that of high-res result in \emph{Vanilla Multi-Scale}. Although the inference speed can be further accelerated with $\gamma$$=$$0.3$, the \emph{End-to-End} performance drops on Text Perceptron and MANGO are much larger than that on Mask-RCNN-based. Since DLD does not involve distillation loss in the text detection task, it is predictable that the \emph{End-to-End} performance can be further improved with the help of the text detection knowledge transfer. Recall the results reported in Table~\ref{tb:1}, and we can find that the FLOPS is negatively correlated to the FPS but is not in equal proportion, because many operations are optimized to calculate in parallel and different platforms might behave differently. Nevertheless, reducing the computational cost is also vital for the low-end devices.
\section{Conclusion}
This paper proposes a novel \emph{Dynamic Low-resolution Distillation} (DLD) framework for cost-efficient end-to-end text spotting, aiming to recognize images in different \emph{low but recognizable} resolutions. The model integrates a dynamic low-resolution selector that can choose different down-sampled scales. A sequential knowledge distillation strategy is then adopted to make the model be able to recognize images with lower resolutions and hence achieve a better resolution-performance balance. Experiments show that our method can effectively enhance the practicability of end-to-end text spotters in many complicated situations.

\newpage

%
%
\bibliographystyle{splncs04}
\bibliography{egbib}

\appendix
\setcounter{table}{0}  
\setcounter{figure}{0}
\renewcommand{\thetable}{A\arabic{table}}
\renewcommand{\thefigure}{A\arabic{figure}}

\newpage

\section{Appendices}

\subsection{More Analysis}
In this section, we will provide more statistical analysis and show some visualization results to demonstrate the effectiveness of the proposed method.

\subsubsection{Selected Scales Distribution.} 
We counted the distributions of the input scales selected by DRS under the entire DLD framework, as illustrated in Table~\ref{tb:2}. As we can see, the model dynamically chooses different input scales for different images. When $\gamma$$=$$0.1$, the model is more inclined to accuracy, and more selections are concentrated on the scales of 0.6, 0.7, or 0.8. With the $\gamma$ increases, the model will select more low-res scales. The scales distributions are also slightly different for different datasets. For example, IC15 chooses more large scales than IC13 since it contains a larger proportion of small text.

\begin{table}
\begin{minipage}[t]{1\linewidth}
\caption{The distributions of the selected scales using different $\gamma$ in DLD.}
\begin{center}
\scalebox{0.85}{
\begin{tabular}{c|c|c|c|c|c|c|c|c|c}
\hline
\multirow{2}{*}{Scales} & \multicolumn{3}{c|}{$\gamma=0.1$} &  \multicolumn{3}{c|}{$\gamma=0.3$} & \multicolumn{3}{c}{$\gamma=0.5$}\\ \cline{2-10}
& IC13  & IC15  & TT & IC13  & IC15  & TT &IC13  & IC15  & TT\\
\hline
0.3 & 0.4\%  & 0.2\%  & 0\% &  4.3\%  & 1.6\%  & 1.6\%  & 17.5\% & 11.4\% & 15.0\%\\
0.4 & 2.6\%  & 0.6\%  & 6.7\% & 15.9\%  & 14.6\%  & 11.3\% & 36.0\% & 35.2\% & 35.3\% \\
0.5 & 7.7\%  & 6.2\%  & 7.3\% & 46.5\%  & 44.2\%  & 47.7\% & 42.3\% & 44.4\% & 42.7\%\\
0.6 & 24.1\%  & 21.0\% & 26.3\%  & 21.1\%  & 29.2\%  & 28.7\%&  3.8\%&8.6\% & 7.0\%\\
0.7 & 34.8\%  & 39.3\% & 38.3\% & 12.0\%  & 10.0\%  & 10.7\% & 0.4\% & 0.4\% & 0\%\\
0.8 & 30.4\%  & 32.7\% & 27.4\% & 0.4\%  & 0.4\%  & 0\%& 0\%&  0\%& 0\%  \\
\hline
\end{tabular}
}
\end{center}
\label{tb:2}
\end{minipage}
\end{table}

\subsubsection{Visualization of the DRS.} 
\begin{figure}[ht]
\begin{center}
\includegraphics[width=0.8\linewidth]{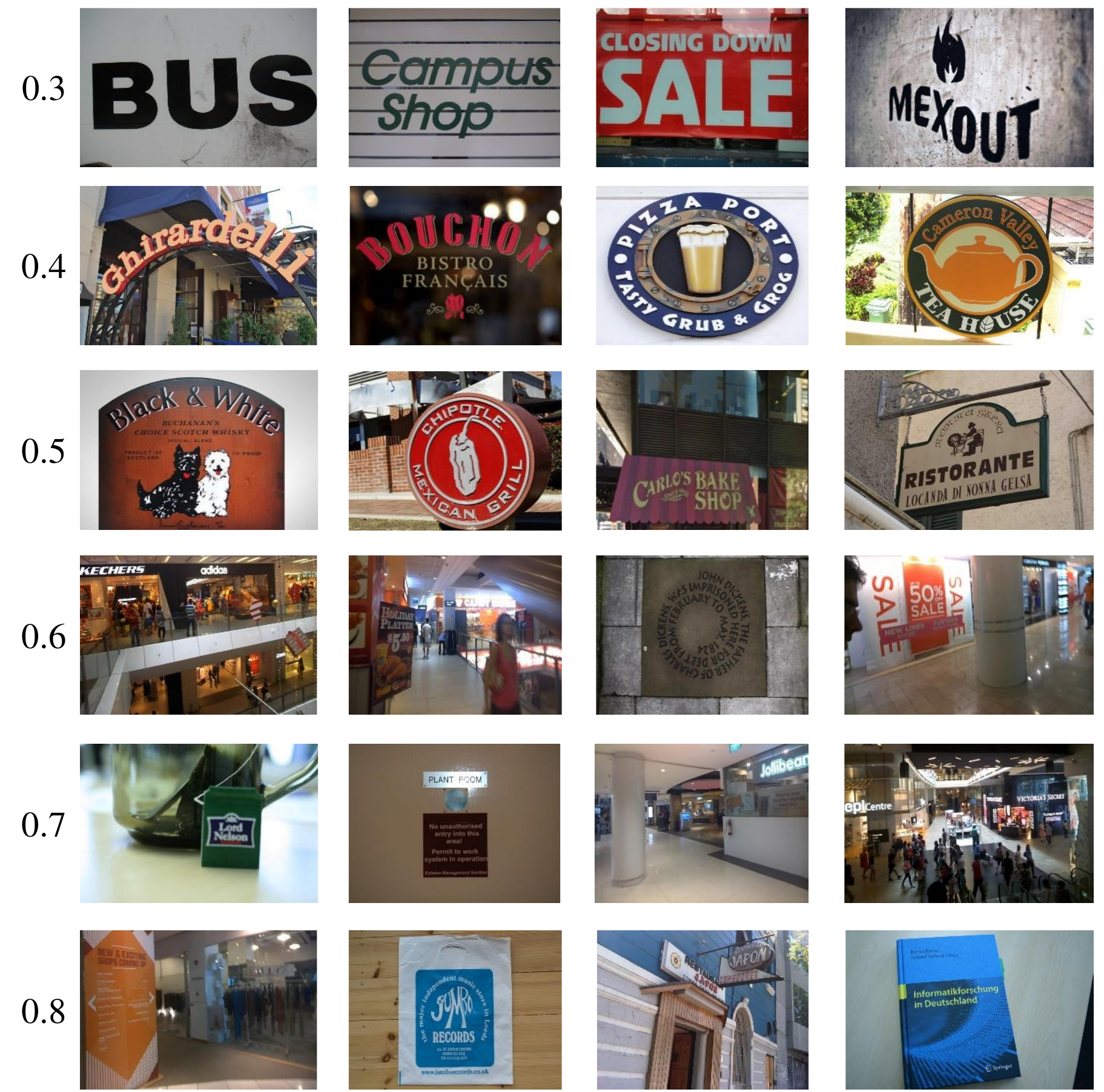}\\
\end{center}
\caption{
The selection results of \emph{DLD-DRS-only} for different scales in IC13, IC15, TT.
}
\label{fig:sup_1}
\end{figure}

Dynamic Resolution Selector (DRS) aims to assign different images with different input scales dynamically. Under the candidate scales set $\{0.3, 0.4, 0.5, 0.6, 0.7, 0.8\}$, we randomly select some of the images from three benchmarks in different scales and visualize them in Figure~\ref{fig:sup_1}. On the overall trend, we can see that DRS is optimized to choose smaller input scales for images containing large text and bigger input scales for images with small text.  Here, the term `large' or `small' reflects the relative ratio of the text to the whole image instead of the instance's absolute resolution. The scale that benefits the overall performance would be selected for images containing both large and small text, such as the sample in row-3 and column-1.

\subsubsection{Visualization of the SKD.}  
\begin{figure*}[t]
\begin{center}
\includegraphics[width=1.0\linewidth]{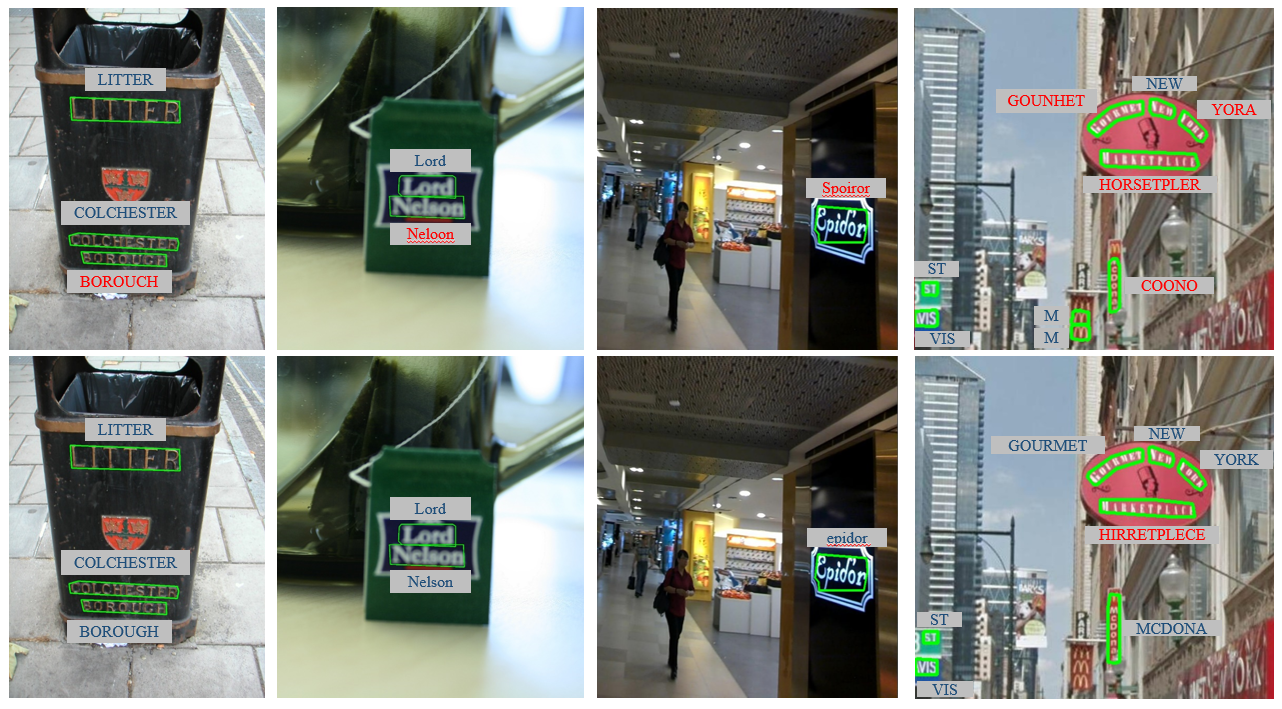}\\
\end{center}
\caption{
The visualization of end-to-end results in IC13, IC15, and TT under $1/2$ resolution inputs. The first row shows the results without using SKD (\emph{Vanilla Multi-Scale}) and the second row adopts SKD (\emph{DLD-SKD-only}). Text in red are incorrectly recognized.}
\label{fig:sup_2}
\end{figure*}

The proposed Sequential Knowledge Distillation (SKD) strategy could help low-resolution images obtain similar performance to those in high-resolution. In Figure~\ref{fig:sup_2}, we demonstrate some end-to-end results before and after adopting the SKD strategy under the $1/2$ resolution inputs (S-384 for IC13, S-640 for IC15 S-448 for TT), respectively. We can see that with the proposed SKD, the model can correctly recognize some of the confused text through context semantic information.

\subsubsection{Visualization of the DLD.}
\begin{figure*}[t]
\begin{minipage}[t]{1\linewidth}
\begin{center}
\includegraphics[width=1.0\linewidth]{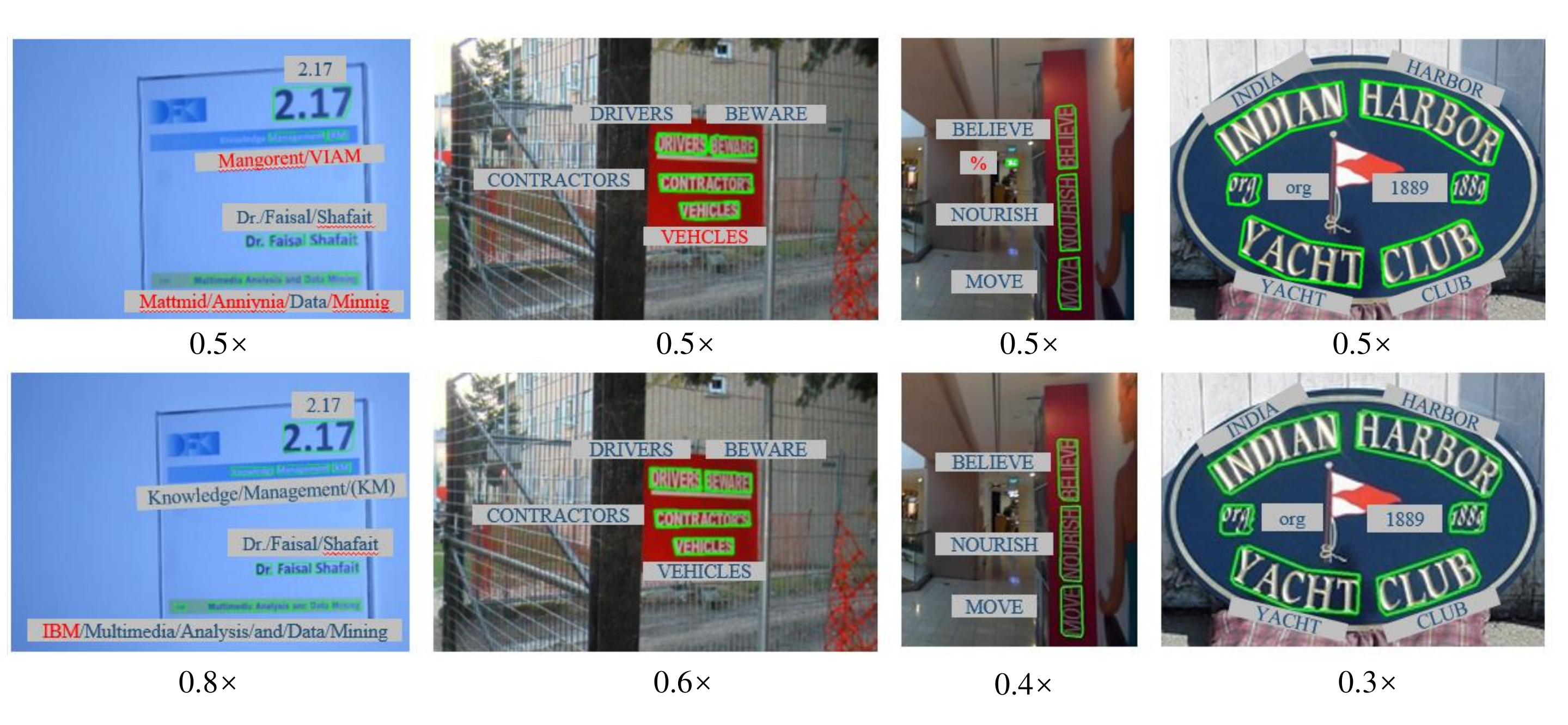}\\
\end{center}
\caption{
The visualization of end-to-end results in IC13, IC15, and TT. The first row shows the results under $1/2$ input scales in \emph{DLD-SKD-only} and the second row corresponds to the entire \emph{DLD}. The numbers below the images are the down-sampled scales compared to the original high-resolution. Text in red are incorrectly recognized.}
\label{fig:sup_3}
\end{minipage}

\begin{minipage}[t]{1\linewidth}
\begin{center}
\includegraphics[width=0.6\linewidth]{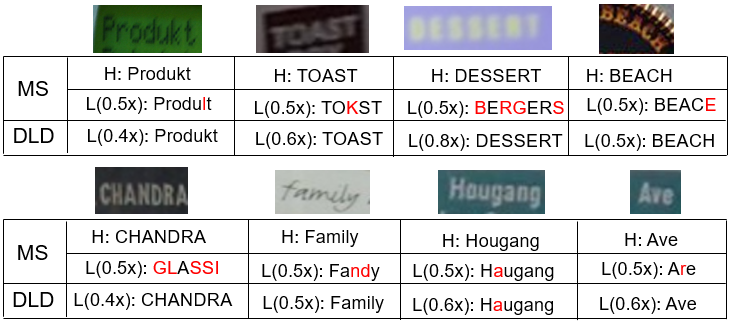}\\
\end{center}
\caption{
The compared recognition results between \emph{Vanilla Multi-Scale} and \emph{DLD}. Characters in red are incorrectly recognized.
}
\label{fig:sup_4}
\end{minipage}
\end{figure*}

Combining with SKD, the model makes the smaller scales have more chances of selecting the DRS module. As for those low-res inputs, DRS can help them select different feasible scales in a cost-efficient way. We illustrate some of the comparisons of End-to-End results between the model of \emph{DLD-SKD-only} and the entire DLD framework in Figure \ref{fig:sup_3}. The model of \emph{DLD-SKD-only} inferences images in the $1/2$ scales. We can see that although some of the blurred text has been correctly recognized, some text still failed to predict since the infeasible fixed scale, such as the first group of images as shown. Under the entire DLD framework, images can be resized to a suitable scale, achieving better performance with minimal computational cost changes. For those easily recognized images such as the fourth group images in Figure \ref{fig:sup_3}, the DLD allows them to choose a lower input scale without performance reduction.

We select some of the samples that are corrected recognized in high-res but failed in low-res by \emph{Vanilla Multi-Scale} model, as visualized in Figure~\ref{fig:sup_4}.
We also provide the compared results predicted by our proposed DLD model.
As we can see, under DLD framework, the model has dynamically selected different scales and mostly correctly recognized these instances in the low resolution.

\begin{figure*}[t]
\begin{center}
\includegraphics[width=0.95\linewidth]{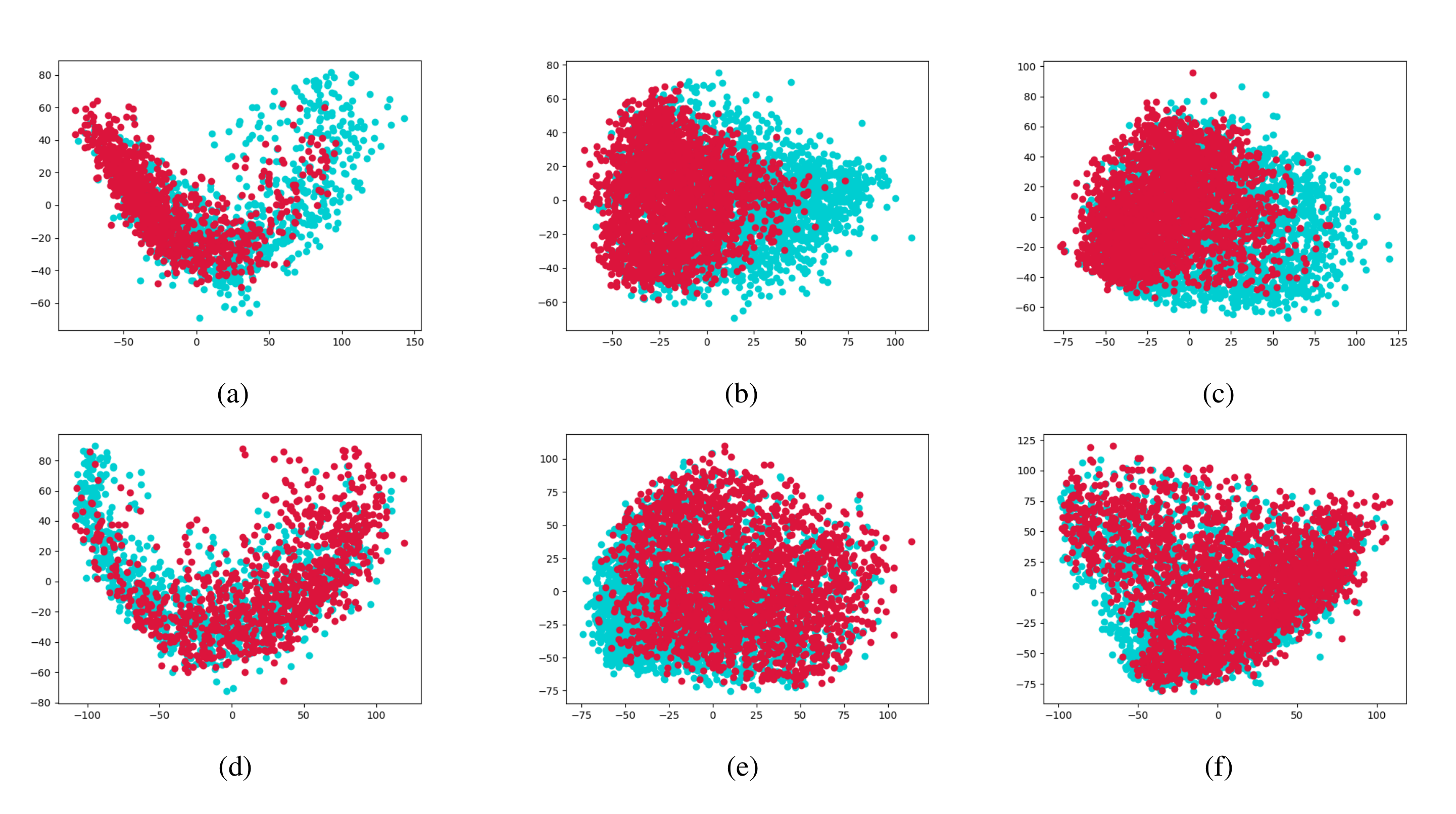}\\
\end{center}
\caption{
The PCA visualization of the \emph{RoI Feature} distributions on high-res and low-res inputs. (a)  \emph{Vanilla Multi-Scale} on IC13. (b) \emph{Vanilla Multi-Scale} on IC15. (c)  \emph{Vanilla Multi-Scale} on TT. (d) \emph{DLD-SKD-only} on IC13. (e) \emph{DLD-SKD-only} on IC15. (f)  \emph{DLD-SKD-only} on TT. Blue points denote the high resolution distribution and Red points denote the low resolution distribution. }
\label{fig:sup_5}
\end{figure*}
\begin{figure*}
\begin{center}
\includegraphics[width=0.95\linewidth]{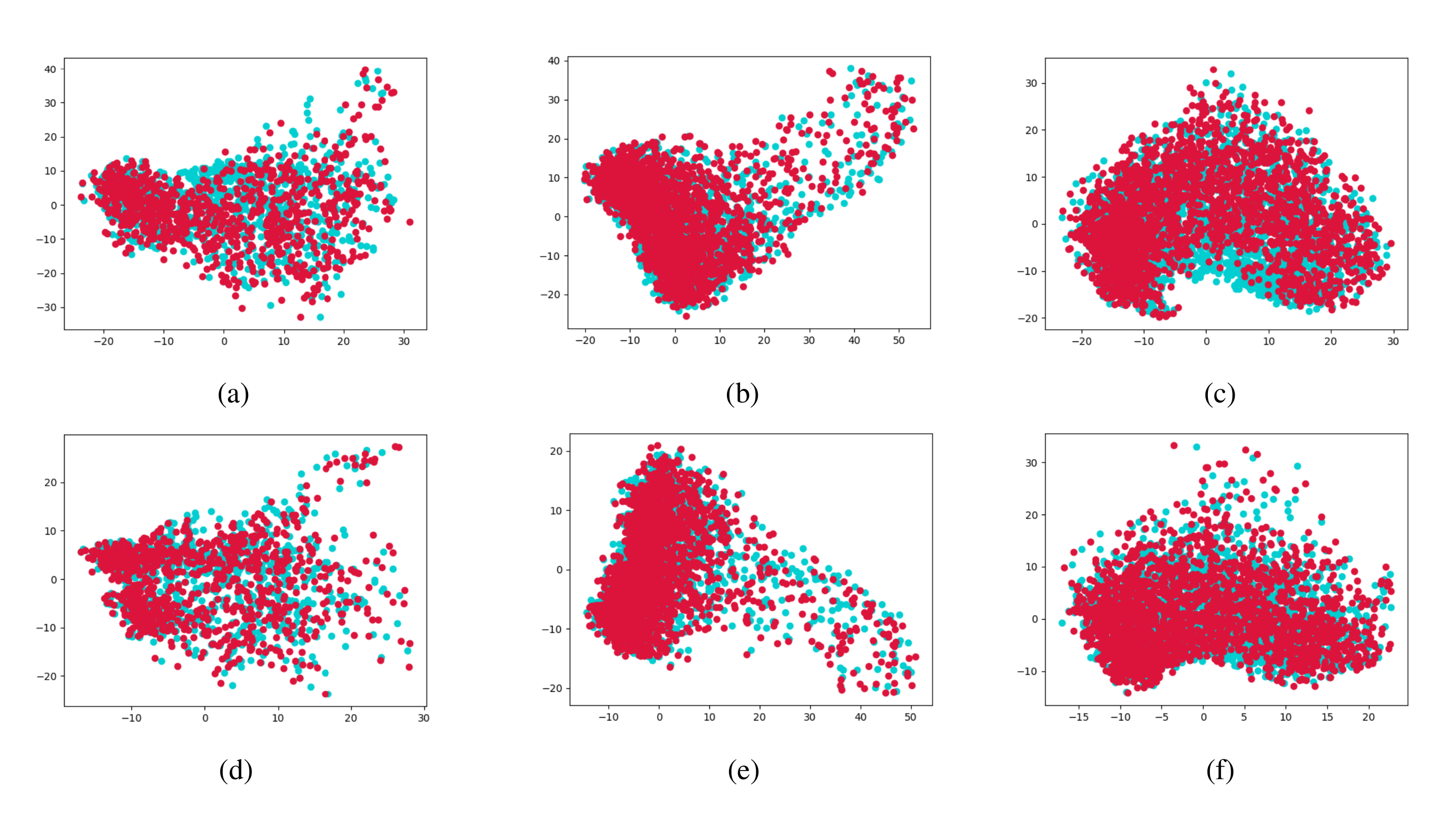}\\
\end{center}
\caption{
The PCA visualization of the \emph{Contexture Feature} distributions on high-res and low-res inputs. (a) \emph{Vanilla Multi-Scale} on IC13. (b) \emph{Vanilla Multi-Scale} on IC15. (c) \emph{Vanilla Multi-Scale} on TT. (d) \emph{DLD-SKD-only} on IC13. (e) \emph{DLD-SKD-only} on IC15. (f) \emph{DLD-SKD-only} on TT. Blue points denote the high resolution distribution and Red points denote the low resolution distribution.  }
\label{fig:sup_6}
\end{figure*}

\subsubsection{Feature Distribution of SKD.}
The target of the SKD is to narrow the feature distribution difference between high resolution and low-resolution inputs. To demonstrate the effectiveness, we visualize the feature distribution in two network positions obtained by Principal Component Analysis (PCA), \emph{i.e.,} (1) \emph{RoI Feature}: the feature map after RoI-Masking operation, and (2) \emph{Contexture Feature}: the feature map after Bi-LSTM module. We compare the feature distribution between high-res and low-res inputs for different models, and the visualization results on three datasets are shown in Figure \ref{fig:sup_5} and Figure \ref{fig:sup_6}.
As we can see, without using the SKD strategy (\emph{Vanilla Multi-Scale}), the feature differences between high-res and low-res inputs are relatively large. When the model integrates with the SKD (\emph{DLD-SKD-only}), the feature distributions are narrowed. This demonstrates that, although the Multi-Scale training strategy improves the models' robustness on different input scales, the model actually assigns different scales with different feature distributions and results in performance discrepancies when input scales change. The SKD strategy essentially aligns the feature distribution differences and makes the model more robust in low-resolution.

Table \ref{tb:7} records the quantitative differences calculated by L2 distance. Compared with \emph{Vanilla Multi-Scale} training, distillation training can reduce the L2 distance of \emph{RoI Feature} distribution by 0.13 on IC13, 0.11 on IC15 and 0.13 on Total-Text, respectively. As for the \emph{Contextual Feature}, the distances can be decreased by 0.08, 0.08, 0.10, respectively.
Comparing the feature discrepancies in two positions, the RoI feature's differences between high-res and low-res inputs are much larger than the other. This demonstrates that it is unwise only to adopt distillation in the deepest layer. Distillation on RoI features can help the model recover the information in the early phases and prevent over-fitting.

\begin{table}[t]
\begin{center}
\scalebox{0.8}{
\begin{tabular}{c|c|c|c}
\hline
Dataset & Training Method &  \makecell[c]{L2 distance of \\ RoI Feature} & \makecell[c]{L2 distance of \\ Contexture Feature}\\
\hline
\multirow{2}{*}{IC13} & Vanilla Multi-Scale & 0.43 & 0.20 \\
 & DLD-SKD-only & 0.30 & 0.12 \\
 \hline
 \multirow{2}{*}{IC15} & Vanilla Multi-Scale & 0.28 & 0.18 \\
 & DLD-SKD-only & 0.17 & 0.10 \\
 \hline
 \multirow{2}{*}{TT} & Vanilla Multi-Scale & 0.34 & 0.19 \\
 & DLD-SKD-only & 0.21 & 0.09 \\
\hline
\end{tabular}
}
\end{center}
\caption{L2 distances of feature distribution between high-res and low-res inputs on different benchmarks.}
\label{tb:7}

\end{table}

\subsection{Discussion}

The technique of End-to-End text spotting has been studying for several years since \cite{DBLP:conf/iccv/LiWS17} was proposed. Many advances have demonstrated the advantages of End-to-End models compared to the traditional two-staged pipeline, for example, (1) better performance since freeing from error accumulation between two tasks, (2) faster inference and smaller storage requirements by information sharing and jointly optimization
(3) lower maintenance cost \cite{DBLP:journals/csur/ChenJZLW21}.

However, most of the current Optical Character Recognition (OCR) engines still solve the real tasks in a two-staged way, such as Tesseract~\cite{DBLP:conf/icdar/Smith07}, PP-OCR~\cite{DBLP:journals/corr/abs-2009-09941}, Calamari~\cite{DBLP:journals/corr/abs-1807-02004}, \emph{etc}.
There might exist many reasons, but one of the immediate factors is that
an end-to-end model with good performance and robustness is hard to obtain in many real complicated situations.
Due to the different characteristics of the sub-tasks, it is not easy to balance two tasks \cite{DBLP:journals/corr/abs-2110-10405}. The text detection task predicts instances' scopes and locations and focuses more on the different scales of coarse-grained features like boundaries. The recognition task is a sequential classification problem that usually requires images on a uniform scale, and it is more concentrated on fine-grained features like textures.  Therefore, the input resolution becomes a sensitive factor to the second staged task and may cause the overall performance fluctuation.

To alleviate such problems, besides the intuitive way that makes the model be trained on more samples and with abundant means of augmentations, the proposed DLD framework can effectively reduce the resolution sensitivity of the text recognition task from the feature level. In real applications, the DLD can be flexibly used to enhance the robustness of the end-to-end model.

\subsubsection{Failure cases.} 
\begin{figure*}
\begin{center}
\includegraphics[width=0.95\linewidth]{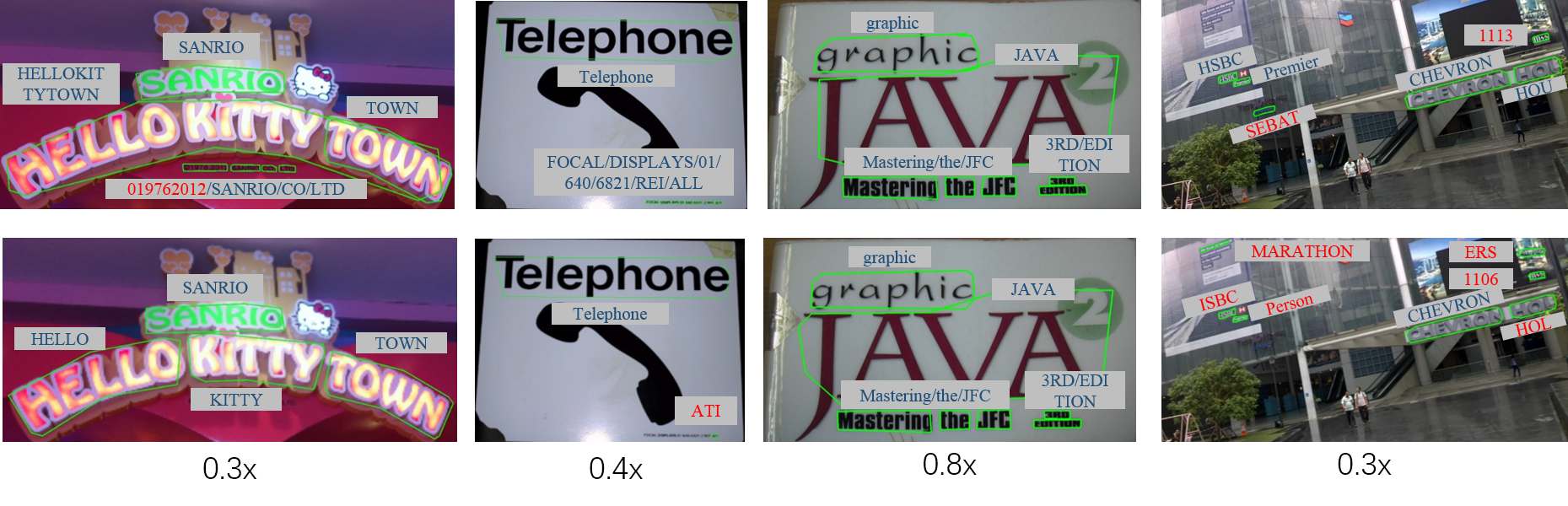}\\
\end{center}
\caption{
Visualization of some failure cases caused by wrong resolution selection in IC13, IC15, and TT. The first row shows the results under original input scales with \emph{Vanilla Multi-Scale} and the second row corresponds to the entire \emph{DLD}. The numbers below the images are the down-sampled scales compared to the original high-resolution. Text in red are incorrectly recognized.
}
\label{fig:sup_7}
\end{figure*}

In this section, we demonstrate some failure cases in the testing phase. The failure cases are mainly caused by wrong resolution selection and errors inherent in the detection model, here we only focus on the failure cases caused by the wrong resolution selection to reveal the limitations of our proposed framework. As shown in \ref{fig:sup_7}, we can see that it is difficult for the resolution selector to choose a suitable resolution when the text size distribution in the images varies greatly. Moreover, there are also a few failure cases assigning large resolutions to images with large text or assigning small resolutions to images with small text, which will brings extra time consumption or makes the text unrecognizable. Nonetheless, our proposed resolution selector works well for most text images and reduces the input resolution without affecting the performance. To demonstrate the effectiveness, as shown in Table~\ref{tb:8}, we illustrate the distribution of failure cases. We can see from the table that most of the failure cases are caused by errors inherent in the detection model. Moreover, errors due to the introduction of our proposed \emph{DLD} framework only account for 2.51$\%$ in IC13, 2.16$\%$ in IC15 and 1.94$\%$ in TT, respectively. Furthermore, equipped with our proposed sequential distillation, we can effectively correct some of incorrect recognition results.
\begin{table}[t]
\begin{center}
\scalebox{0.8}{
\begin{tabular}{c|c|c|c}
\hline
Dataset & \emph{Vanilla Multi-Scale} &  \emph{DLD} & ratio ($\%$)\\
\hline
\multirow{4}{*}{IC13} & \ding{52} & \ding{52} & 78.08 \\
& \ding{52} & \ding{55} & 2.51 \\
& \ding{55} & \ding{52} & 3.49 \\
& \ding{55} & \ding{55} & 15.92 \\
 \hline
\multirow{4}{*}{IC15} & \ding{52} & \ding{52} & 62.93 \\
& \ding{52} & \ding{55} & 2.16 \\
& \ding{55} & \ding{52} & 4.72 \\
& \ding{55} & \ding{55} & 30.19 \\
 \hline
\multirow{4}{*}{TT} & \ding{52} & \ding{52} & 64.79 \\
& \ding{52} & \ding{55} & 1.94 \\
& \ding{55} & \ding{52} & 4.74 \\
& \ding{55} & \ding{55} & 28.53 \\
\hline
\end{tabular}
}
\end{center}
\caption{The distribution of failure cases in IC13, IC15 and TT. $\ding{52}$ means correct text recognition, $\ding{55}$ means incorrect text recognition.}
\label{tb:8}
\end{table}

\subsubsection{Limitations.} 
Although the proposed DLD framework effectively optimizes the model's performance of accuracy and computational cost, there are still some problems to be solved in the future.

First, the basic high-resolution and the set of candidate down-sampled scales need to be decided manually, which requires sufficient data analysis and certain experiences of people. Given an empirical setting, although the model can perform better than simply using multi-scale training and fixed-scale testing, but can hardly find out the theoretically best solution. Enlarging the candidate set's capacity is a way to find more optimal results. However, the training cost will inevitably increase since the selector needs to calculate the forward results of all candidate scales during training.  A possible alternative way to achieve dynamic resolution selecting is to predict resolutions in soft labels rather than hard labels, such as quality assessment~\cite{DBLP:journals/ijcm/HeGHH14}. However, we still face many challenges since it is hard to define the quality of an image containing different text.

Second, Although the KD strategy can increase the performance of the low-res model, there is always a limit to the competence of the model. When the input image is down-sampled to a small scale that even humans cannot recognize, the KD learning might lead the model to an over-fitting state.
Furthermore, the current model only considers the KD problem on the recognition task. However, the detection task is also sensitive to the resolution for the text spotters that adopt the segmentation-based detection branch. Although we have demonstrated the effectiveness to adopt our method on those models, the performances still have large improvement spaces. More experiments should be conducted on the different types of models in the future.

The last problem is a common problem for most current text spotters. The target of text spotting is to obtain the final text sequences prediction. However, the text detection branch is mostly optimized to obtain high-IoU with the detection Ground Truth (GT), which is not always consistent with text recognition~\cite{DBLP:conf/aaai/QiaoCCXNPW21,DBLP:journals/corr/abs-2110-10405}. This problem would introduce the new question about how to balance these two tasks. In our DLD, in addition to the GT-oriented optimization, many other balance parameters still need to be set manually, which might also influence the model's performance to some extent.

\end{document}